\pdfoutput=1

\documentclass[11pt]{article}

\usepackage{EMNLP2022}

\usepackage{times}
\usepackage{latexsym}

\usepackage[T1]{fontenc}

\usepackage[utf8]{inputenc}

\usepackage{microtype}

\usepackage{inconsolata}

\usepackage{graphicx}
\usepackage{subfigure}
\usepackage{enumitem}
\usepackage{amsmath}
\usepackage{amssymb}
\usepackage{bbm}
\usepackage{booktabs}
\usepackage{multirow}
\usepackage{makecell}
\usepackage{pifont}

\newcommand{\ct}{\textsf{\small CLS}-tuning}
\newcommand{\pt}{\textsf{\small Prompt}-tuning}
\usepackage{CJKutf8}
\newcommand{\zh}[1]{\begin{CJK}{UTF8}{gbsn}#1\end{CJK}}

\newcommand{\coconut}{\includegraphics[width=0.02\textwidth]{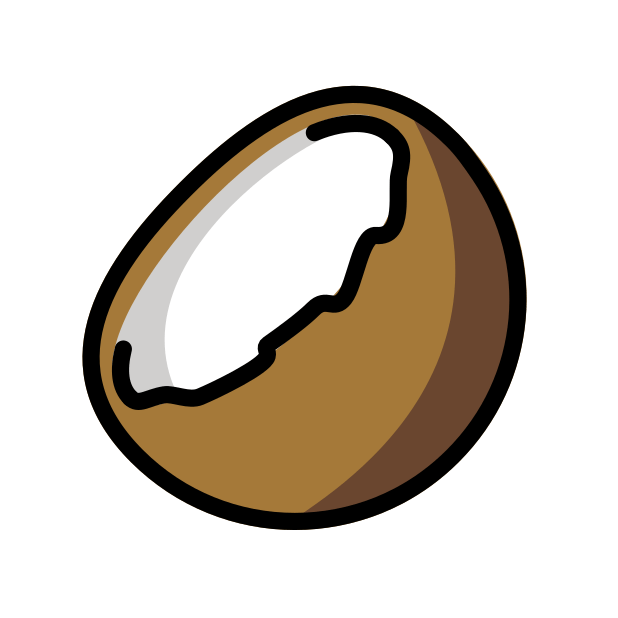}}
\newcommand{\kiwi}{\includegraphics[width=0.02\textwidth]{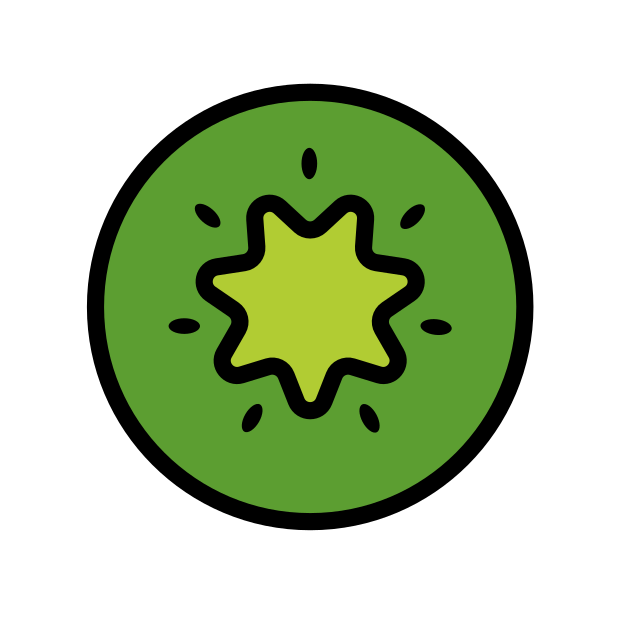}}
\newcommand{\lemon}{\includegraphics[width=0.02\textwidth]{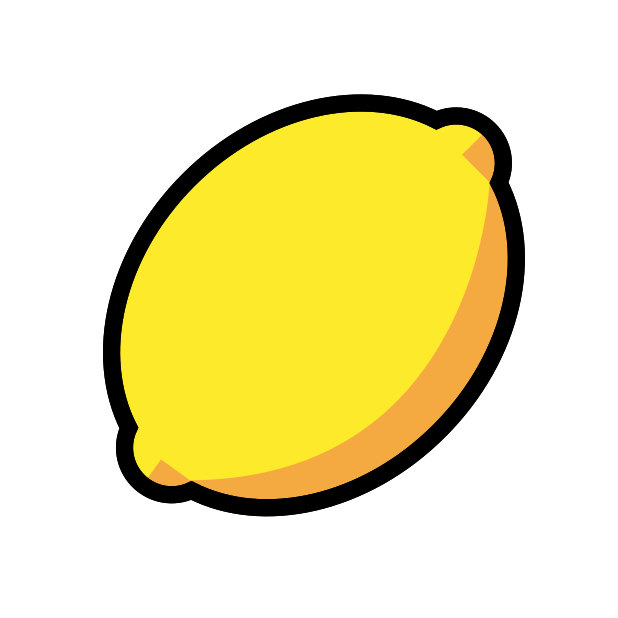}}
\newcommand{\glee}{\includegraphics[width=0.02\textwidth]{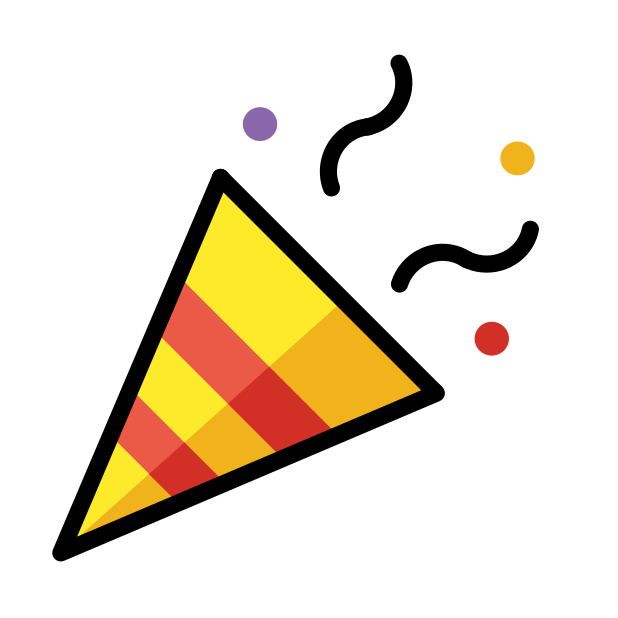}}

\title{Making Pretrained Language Models Good Long-tailed Learners}

\author{Chen Zhang\textsuperscript{\coconut}, Lei Ren\textsuperscript{\kiwi}, Jingang Wang\textsuperscript{\kiwi\lemon}, Wei Wu\textsuperscript{\kiwi}, Dawei Song\textsuperscript{\coconut\lemon}\Thanks{\textsuperscript{\lemon}Jingang Wang and Dawei Song are the corresponding authors.} \\
\textsuperscript{\coconut}Beijing Institute of Technology \\
\texttt{\{czhang,dwsong\}@bit.edu.cn} \\
\textsuperscript{\kiwi}Meituan NLP \\
\texttt{\{wangjingang02,wuwei30\}@meituan.com} \\
\texttt{renlei\_work@163.com} \\}

\makeatletter
\def\thanks#1{\protected@xdef\@thanks{\@thanks
    \protect\footnotetext{#1}}}
\makeatother

\begin{document}

\maketitle

\begin{abstract}
Prompt-tuning has shown appealing performance in few-shot classification by virtue of its capability in effectively exploiting pre-trained knowledge. This motivates us to check the hypothesis that prompt-tuning is also a promising choice for long-tailed classification, since the tail classes are intuitively few-shot ones. To achieve this aim, we conduct empirical studies to examine the hypothesis. The results demonstrate that prompt-tuning makes pretrained language models at least good long-tailed learners. For intuitions on why prompt-tuning can achieve good performance in long-tailed classification, we carry out in-depth analyses by progressively bridging the gap between prompt-tuning and commonly used finetuning. The summary is that the classifier structure and parameterization form the key to making good long-tailed learners, in comparison with the less important input structure. Finally, we verify the applicability of our finding to few-shot classification.\footnote{\underline{G}ood \underline{l}ong-tailed l\underline{e}arn\underline{e}rs can be abbreviated as \glee\textsc{Glee}. Code and data are available at \url{https://github.com/GeneZC/Glee}.} 
\end{abstract}

\section{Introduction}

\label{sec1}

\begin{figure}
    \centering
    \includegraphics[width=0.37\textwidth]{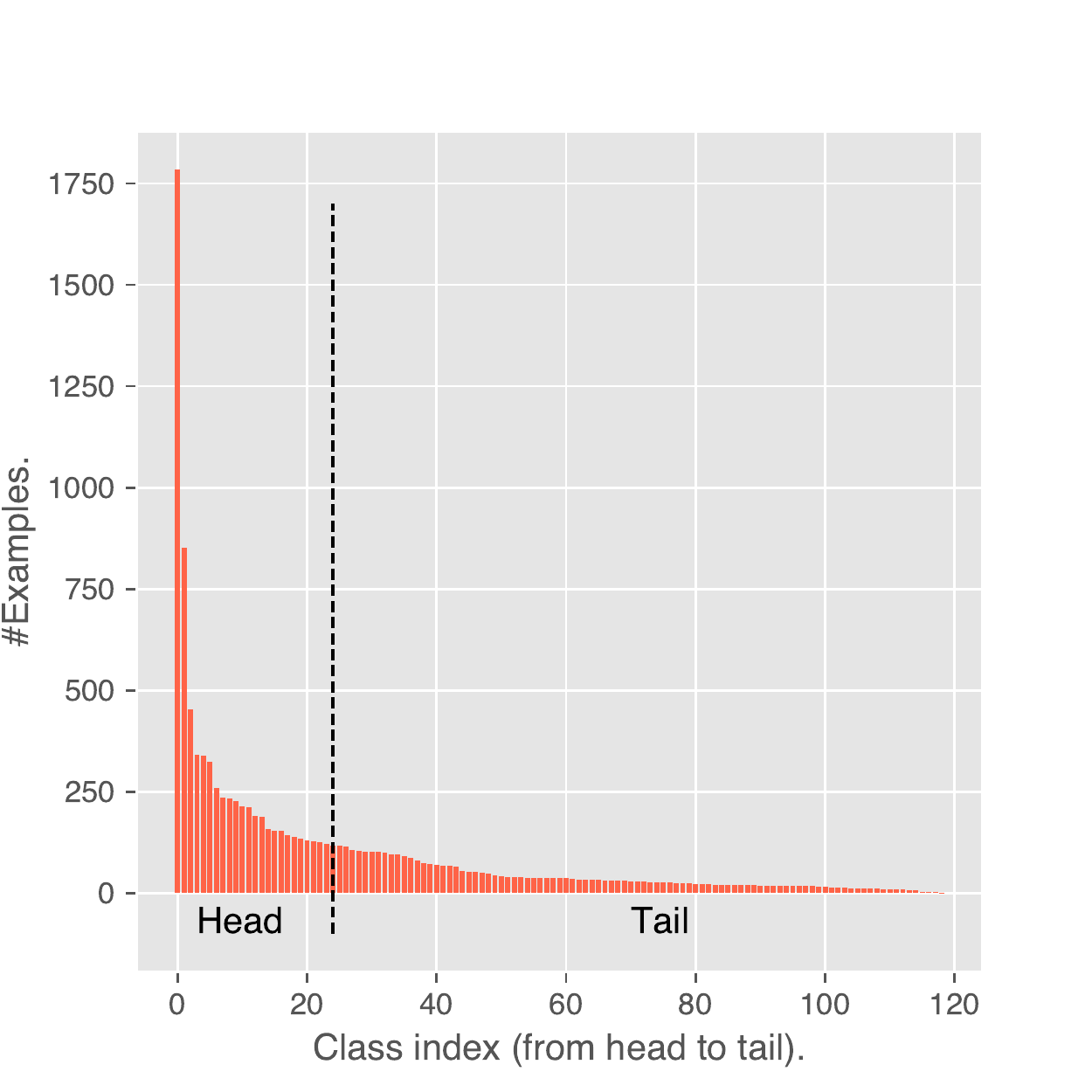}
    \caption{An example long-tailed class distribution from \textsc{Iflytek} dataset~\citep{Xu20}, where we can distinguish tail classes from head classes.}
    \label{fig1}
\end{figure}

Pretrained language models (PLMs) with \ct~(i.e., finetuning a PLM by applying a classifier head over the \texttt{[CLS]} representation) have achieved strong performance in a wide range of downstream classification tasks~\citep{Devlin19,Liu19,Raffel20,Wang19a}. However, they have been less promising in the long-tailed scenario~\citep{Li20}. The long-tailed scenario is different from common scenarios due to the long tail phenomenon exhibited in the class distribution, as illustrated in Figure~\ref{fig1}. The long-tailed class distribution prevents PLMs from achieving good performance, especially in tail classes that only allow learning with very few examples (dubbed tail bottleneck). 

Recent advances on \pt\footnote{\pt~can be an ambiguous term regarding the parameter-efficient prompt tuning~\citep{Lester21}. However, we insist on the use of it to accord with \ct.}~have witnessed a surge of making PLMs better few-shot learners~\citep{SchickS21,Gao21,Scao21}. \pt~requires PLMs to perform classification in a cloze style, thus is superior to \ct~in two aspects: 1) it aligns the input structure with that of masked language modeling (MLM); and 2) it re-uses the classifier structure and parameterization from the pretrained MLM head. These two merits of \pt~equip PLMs to better exploit pretrained knowledge and hence gain better few-shot performance than that with \ct.

The success of \pt~in few-shot scenarios motivates us to hypothesize that \pt~can relieve the tail bottleneck and thus make PLMs good long-tailed learners. The reason why we make such a hypothesis is that the tail classes are intuitively few-shot ones. However, long-tailed classification is different from few-shot classification to a certain extent, as it allows the possibility to transfer knowledge from head classes to tail ones. 

We empirically examine the hypothesis by conducting empirical evaluations on three long-tailed classification datasets. The comparison results show that PLMs can be good long-tailed learners with \pt, which outperforms PLMs with \ct~by large margins. Besides, \pt~even exhibits better performance than that of \ct~with appropriate calibrations (e.g., focal loss~\citealt{Lin17}). The widely accepted decoupling property~\citep{Kang20} claims that a good long-tailed learner should enjoy a nearly uniform distribution in terms of weight norms, otherwise the norms of head classes can be way much larger than the tail ones'. It is therefore expected that the weights tuned with \pt~own a flat distribution. We validate the property of \pt~by visualizing the norms of trained classification weights across classes. With the compelling results, we put that our hypothesis is valid.

We also provide further intuitions by asking why \pt~could be so promising, as shown in the above empirical investigations. Through in-depth analyses, we uncover that re-using the classifier structure and parameterization from the MLM head is a key for attaining the good long-tailed performance, largely outweighing the importance of aligning the input structure with that of MLM. \ct, with classifier structure derived and parameters partly initialized from the MLM head, approximates the performance of \pt. We believe that this observation would as well shed light on related work that aims to improve \pt~itself. As such, we finally present comparison results of the improved \ct~and \pt~in the few-shot scenario, suggesting the applicability of the improved \ct~to few-shot classification.

\section{Background}

\subsection{Long-tailed Classification}

Long-tailed classification basically follows a classification setting. Given a dataset $\mathcal{D}=\{(x_i,y_i)\}_i$ in which $(x,y)\sim P(\mathcal{X},\mathcal{Y})$, a model $\mathcal{M}$ is required to learn to approximate $P(\mathcal{Y}\mid \mathcal{X})$ as accurate as possible so that it can correctly predict the label from the input. However, the long-tailed classification differs from the common classification setting in that $P(\mathcal{Y})$ is a long-tailed one, prohibiting $\mathcal{M}$ from achieving a good optimization, especially on tail classes.

\subsection{Finetuning}

\begin{figure*}
    \centering
    \subfigure[\ct]{
    \includegraphics[width=0.37\textwidth]{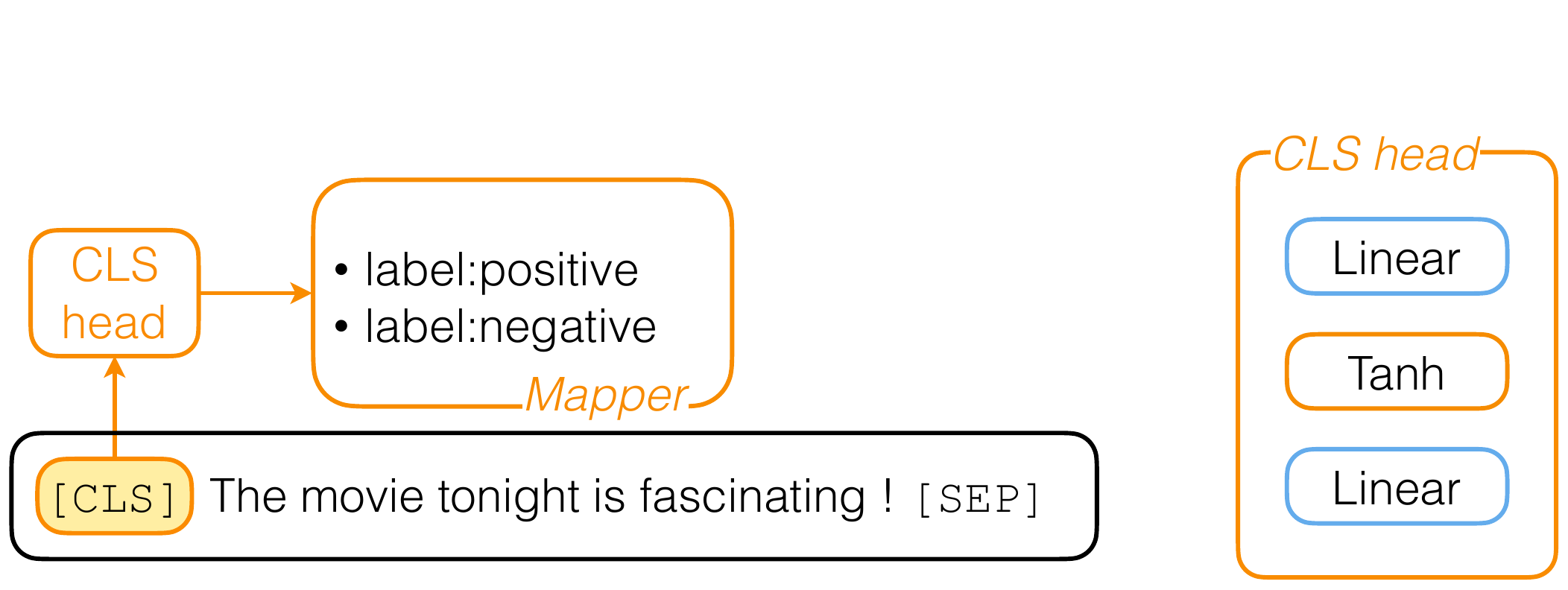}
    \label{figta}
    }
    \subfigure[\pt]{
    \includegraphics[width=0.47\textwidth]{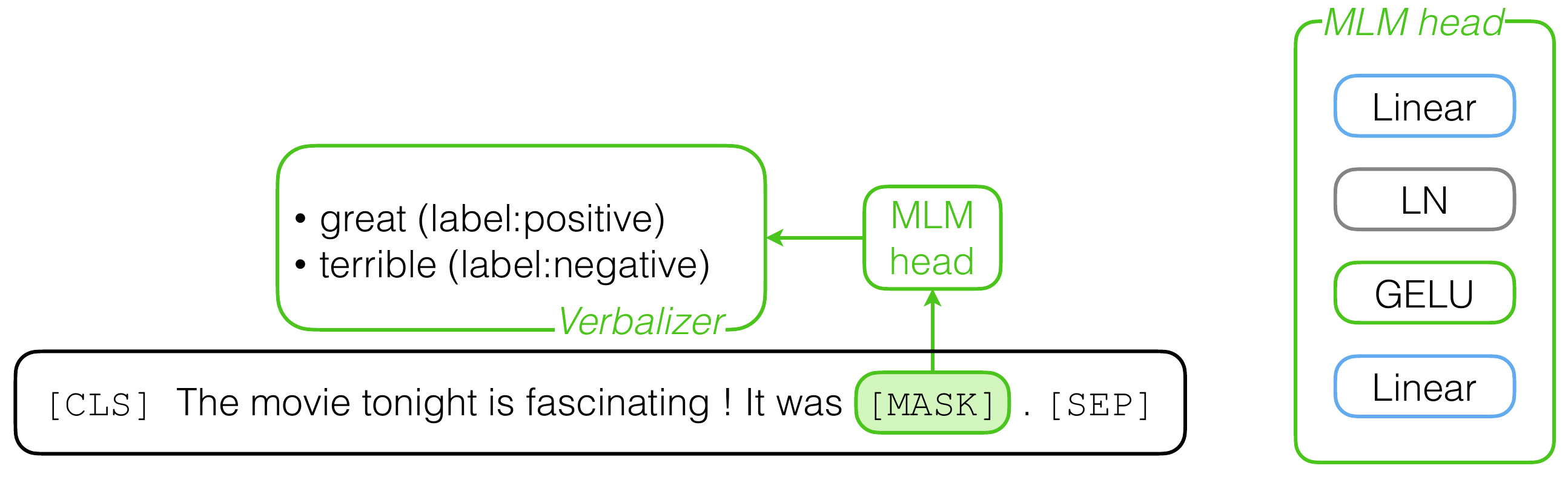}
    \label{figtb}
    }
    \caption{Illustration of two finetuning schemes.}
    \label{figt}
\end{figure*}

\paragraph{CLS-tuning}

Pretrained with the special token \texttt{[CLS]} for overall semantics, PLMs can be finetuned with classifiers over the \texttt{[CLS]} representations for classification (Figure~\ref{figta} left).  

The optimization objective can be depicted as:
\begin{gather}
    \mathbb{L}_{\textsf{CLS}}=-\log P(y_i\mid x_i;\mathcal{M}),
\end{gather}
which is exactly a cross entropy loss. Here, $\mathcal{M}$ can be disassembled to a backbone $\mathcal{E}$ and a classifier $\mathcal{C}$. While $\mathcal{E}$ is a PLM producing \texttt{[CLS]} representation, $\mathcal{C}$ is a Tanh-activated MLP. Here, the MLP typically consists of two feed-forward linear layers. To be more specific, we name the classifier for \ct~as CLS head (Figure~\ref{figta} right), and generally the last layer of the classifier as predictor. We also name the input as CLS input for brevity.

\paragraph{Prompt-tuning}

For PLMs that are pretrained with an MLM objective, it is natural to finetune the PLMs in an MLM-like cloze style for better exploitation of the pretrained knowledge. 

To reach the goal, a template $\mathcal{T}$ and a verbalizer $\mathcal{V}$ are introduced~\citep{SchickS21}. The template converts the original input to an input with one \texttt{[MASK]} token that should be recovered, in other words MLM input. The verbalizer maps all labels to their corresponding tokens, and the model should predict the token corresponding to the correct label. In particular, for a label that is mapped to multiple tokens, one \texttt{[MASK]} should be faced with the issue of inability of multi-token completion. We are inspired by the average strategy in~\citet{Chen21,Hu21}, and treat the average of logit values for multiple tokens as the logit value for the label. 

For example (Figure~\ref{figtb} left), the template for sentiment classification can be:
\begin{gather}
    \mathcal{T}(x)=x\text{. It was \texttt{[MASK]}.},
\end{gather}
Accordingly, the verbalizer can be:
\begin{gather}
    \mathcal{V}(y)=
    \begin{cases}
    \text{great} & y\text{ is label:positive} \\
    \text{terrible} & y\text{ is label:negative} \\
    \end{cases},
\end{gather}

Thereby, the optimization objective is described as:
\begin{gather}
    \mathbb{L}_{\textsf{Prompt}}=-\log P(\mathcal{V}(y_i)\mid \mathcal{T}(x_i);\mathcal{M}),
\end{gather}
where $\mathcal{E}$ in $\mathcal{M}$ generates the \texttt{[MASK]} representation, and $\mathcal{C}$ is the pretrained MLM head. The MLM head (Figure~\ref{figtb} right) is activated with a GELU~\citep{Hendrycks16} and normalized with a layer normalization~\citep{Ba16,Vaswani17,Devlin19}. Note that $\mathcal{E}$ and $\mathcal{C}$ share a part of parameters (i.e., the word embeddings in $\mathcal{E}$ and the predictor over the vocabulary in $\mathcal{C}$).

\subsection{Research Hypothesis}

As discussed in the leading Section~\ref{sec1}, we observe that, in a long-tailed class distribution, each of the tail classes is provided with very few examples, typically fewer than one tenth of the number of a common class. This brings challenges in long-tailed classification. Meanwhile, \pt~has been demonstrated to make PLMs better few-shot learners by exploiting pretrained knowledge. 

Therefore, we are inspired to hypothesize that \pt~can make PLMs good long-tailed learners, as pretrained knowledge is intuitively learned from a long-tailed vocabulary. In the following, we present a series of empirical examinations to test whether our hypothesis is valid or not in Section~\ref{sec3}, and why it is so in Section~\ref{sec4}.

\section{Empirical Examination}

\label{sec3}

\subsection{Setup}

\paragraph{Datasets}

\begin{table*}[t]
    \centering
    \caption{Statistics of the long-tailed datasets.}
    \resizebox{0.73\textwidth}{!}{
    \begin{tabular}{cccccc}
    \toprule
    Dataset & \#Train exam. & \#Dev exam. & \#Test exam. & \#Avg. tokens & \#Classes \\
    \midrule
    \textsc{Cmid} & 8,678 & 1,226 & 2,450 & 32.0 & 36 \\
    \textsc{Iflytek} & 10,920 & 1,213 & 2,599 & 289.2 & 119 \\
    \textsc{Ctc} & 20,666 & 2,296 & 7,682 & 27.2 & 44 \\
    \textsc{Msra} & 106,301 & 11,811 & 8,419 & 82.9 & 26 \\
    \textsc{R52} & 5,943 & 617 & 2,570 & 114.8 & 52 \\
    \bottomrule
    \end{tabular}
    }
    \label{tab1}
\end{table*}

We conduct examinations on five long-tailed classification datasets, ranging from Chinese to English ones. The first one is a medical intent question detection dataset (\textsc{Cmid})~\citep{Chen20}. The second one is an application category classification dataset (\textsc{Iflytek}) maintained by CLUE~\citep{Xu20}. The third one is a clinical trial criterion categorization dataset (\textsc{Ctc})~\citep{Zong21}. The fourth one is an entity typing dataset (\textsc{Msra}) originally released as a named entity recognition dataset~\citep{Levow06}.  The last one is a document topic classification dataset (\textsc{R52}) essentially derived from Reuters 21578 dataset~\citep{Debole04}.

For datasets that originally do not include a test set (e.g., \textsc{Iflytek}), we use the development set as test set and randomly take 10\% of the training set as development set. The statistics of these datasets are listed in Table~\ref{tab1}.

\paragraph{Templates and Verbalizers}

For \pt, example templates for the three datasets separately are shown as below:
\begin{itemize}[topsep=1pt,parsep=1pt]
    \item \textsc{Cmid}: $x$? The intent of the question is \texttt{[MASK]}.
    \item \textsc{Iflytek}: $x$. The mentioned application belongs to \texttt{[MASK]}.
    \item \textsc{Ctc}: $x$. The category of the criterion is \texttt{[MASK]}.
    \item \textsc{Msra}: $x$. The $e$ in the sentence is \texttt{[MASK]}.
    \item \textsc{R52}: $x$. This is \texttt{[MASK]}.
\end{itemize}
where $x$ denotes the input, and $e$ denotes the mentioned entity in the sentence offered in \textsc{Msra}. Here, necessary English translations of Chinese templates are used, and the according Chinese templates are listed in Appendix~\ref{app1}.

Since there are many classes for each dataset, we leave the details on verbalizers to Appendix~\ref{app1}. Basically, the verbalizers are deduced from class descriptions after removal of some less meaningful tokens (e.g., punctuations).

\paragraph{Implementation}

Experiments are carried out on an Nvidia Tesla V100. All models are implemented with PyTorch\footnote{\url{https://github.com/pytorch/pytorch}} and Transformers\footnote{\url{https://github.com/huggingface/transformers}} libraries. We initialize models with the Google-released \texttt{bert-base-chinese} and \texttt{bert-base-uncased} checkpoints\footnote{\url{https://github.com/google-research/bert}}. For parameter settings, the batch size is 32, the learning rate is 1e-5, the weight decay is 0, and lastly the gradient norm is constrained to 1. We train the models for 10 epochs with patience of 2 epochs. In order to stabilize the training procedure, we add a linear warm-up for 1 epoch. The maximum sequence length is set according to the dataset, specifically, 64 for \textsc{Cmid}, 512 for \textsc{Iflytek}, 64 for \textsc{Ctc}, 128 for \textsc{Msra}, and 256 for \textsc{R52}.

\paragraph{Metrics}

Since we are more concerned with model performance across different classes, we  use the macro F1 scores as main performance metric. We also offer the  macro F1 scores of head (Head scores) and tail classes (Tail scores) separately to gain a fine-grained understanding of the model performance. To separate head classes from the tail classes, we sort all classes in descending order according to the number of examples within each class. According to the power law, we should get head classes that take up 80\% of all examples. However, we find some tail classes with very limited examples can be included in this manner. So we manually determine the percentage for each dataset, specifically, 55\% for \textsc{Msra} and \textsc{R52}, 65\% for \textsc{Cmid} and \textsc{Iflytek}, and 80\% for \textsc{Ctc}. In addition, we also gather the accuracy scores (Acc scores) for reference. We take average scores over 5 runs as the results, attached with variances.

\subsection{Comparison Results}

\begin{table*}[t]
    \centering
    \caption{Comparison results. \textsc{Avg} denotes average results over all datasets. The best \textsc{Avg} scores are boldfaced. The variances are attached as subscripts.}
    \resizebox{1.0\textwidth}{!}{
    \begin{tabular}{lcccccccccccc}
    \toprule
    Dataset & \multicolumn{2}{c}{\textsc{Cmid}} & \multicolumn{2}{c}{\textsc{Iflytek}} & \multicolumn{2}{c}{\textsc{Ctc}} & \multicolumn{2}{c}{\textsc{Msra}} &  \multicolumn{2}{c}{\textsc{R52}} & \multicolumn{2}{c}{\textsc{Avg}} \\
    \midrule
    Metric & Acc & F1 & Acc & F1 & Acc & F1 & Acc & F1 & Acc & F1 & Acc & F1 \\
    \midrule
    \ct & 51.1\textsubscript{0.4} & 37.3\textsubscript{2.3} & 58.7\textsubscript{0.4} & 33.7\textsubscript{1.6} & 84.6\textsubscript{0.3} & 77.2\textsubscript{2.9} & 99.0\textsubscript{0.1} & 97.5\textsubscript{1.0} & 95.3\textsubscript{0.2} & 67.3\textsubscript{1.3} & 77.7 & 62.6 \\
    \quad w/ $\eta$-norm & 51.1\textsubscript{0.5} &	37.4\textsubscript{2.0} & 59.1\textsubscript{0.3} & 35.7\textsubscript{1.6} & 84.7\textsubscript{0.2} & 77.3\textsubscript{3.1} & 99.0\textsubscript{0.1} & 97.4\textsubscript{0.9} & 95.4\textsubscript{0.3} & 68.9\textsubscript{1.9} & \textbf{77.9} & 63.3 \\
    \quad w/ focal loss & 51.0\textsubscript{0.7} & 42.1\textsubscript{1.3} & 58.8\textsubscript{0.3} & 36.0\textsubscript{1.6} & 84.3\textsubscript{0.4} & 78.5\textsubscript{2.4} & 99.0\textsubscript{0.1} & 96.8\textsubscript{1.2} & 95.7\textsubscript{0.2} & 72.8\textsubscript{2.3} & 77.8 & 65.2 \\
    \midrule
    \pt & 49.3\textsubscript{0.7} & 43.4\textsubscript{0.7} & 61.2\textsubscript{0.6} & 44.4\textsubscript{1.0} & 84.2\textsubscript{0.1} & 80.9\textsubscript{0.1} & 99.1\textsubscript{0.0} & 97.8\textsubscript{0.3} & 95.7\textsubscript{0.1} & 85.3\textsubscript{0.6} & \textbf{77.9} & \textbf{70.4} \\
    \quad w/ focal loss & 48.6\textsubscript{06}	& 42.5\textsubscript{0.6} & 59.7\textsubscript{0.6} & 43.9\textsubscript{0.7} & 83.5\textsubscript{0.6} & 80.2\textsubscript{0.7} & 99.0\textsubscript{0.1} & 97.2\textsubscript{0.7} & 95.5\textsubscript{0.3} & 82.6\textsubscript{2.4} & 77.3 & 69.3 \\
    \midrule
    Metric & Head & Tail & Head & Tail & Head & Tail & Head & Tail & Head & Tail & Head & Tail \\
    \midrule
    \ct & 50.3\textsubscript{1.0} & 34.1\textsubscript{3.0} & 61.8\textsubscript{0.6} & 27.4\textsubscript{1.9} & 87.7\textsubscript{0.2} & 74.1\textsubscript{3.7} & 99.2\textsubscript{0.1} & 97.4\textsubscript{1.1} & 99.0\textsubscript{0.1} & 66.6\textsubscript{1.3} & 79.6 & 59.9 \\
    \quad w/ $\eta$-norm & 50.3\textsubscript{0.9} & 34.3\textsubscript{2.7} & 62.1\textsubscript{0.4} & 29.7\textsubscript{2.0} & 87.8\textsubscript{0.2} & 74.3\textsubscript{4.0} & 99.2\textsubscript{0.1} & 97.3\textsubscript{1.0} & 99.0\textsubscript{0.1} & 68.3\textsubscript{1.9} & \textbf{79.7} & 60.8 \\
    \quad w/ focal loss & 49.8\textsubscript{0.8} & 40.2\textsubscript{1.5} & 62.0\textsubscript{0.4} & 30.2\textsubscript{1.9} & 87.5\textsubscript{0.3} & 75.9\textsubscript{3.1} & 99.3\textsubscript{0.0} & 96.7\textsubscript{1.3} & 99.0\textsubscript{0.0} & 72.3\textsubscript{2.4} & 79.5 & 63.1 \\
    \midrule
    \pt & 48.4\textsubscript{1.0} & 42.2\textsubscript{0.7} & 63.6\textsubscript{0.4} & 40.1\textsubscript{1.2} & 87.4\textsubscript{0.2} & 79.0\textsubscript{0.2} & 99.2\textsubscript{0.1} & 97.7\textsubscript{0.3} & 98.6\textsubscript{0.1} & 85.0\textsubscript{0.6} & 79.4 & \textbf{68.8} \\
    \quad w/ focal loss & 47.1\textsubscript{0.8} & 41.4\textsubscript{0.8} & 62.3\textsubscript{0.6} & 39.8\textsubscript{0.8} & 86.7\textsubscript{0.5} & 78.3\textsubscript{0.7} & 99.3\textsubscript{0.1} & 97.1\textsubscript{0.7} & 98.8\textsubscript{0.1} & 82.3\textsubscript{2.4} & 78.8 & 67.8 \\
    \bottomrule
    \end{tabular}
    }
    \label{tab2}
\end{table*}

In order to examine whether our hypothesis that \pt~can make PLMs good long-tailed learners is valid or not, we first conduct an evaluation on the long-tailed datasets. The results are given in Table~\ref{tab2}.

A key finding from the comparison results is that \pt~outperforms \ct~by large margins across datasets in terms of F1 scores. And \pt~in fact owns performance with lower variances compared with \ct. \pt~even exhibits better F1 scores than calibrated \ct~(e.g., focal loss in our case). Besides, focal loss does not bring further improvement over \pt, implying \pt~is a sufficiently good long-tailed learner. Contrarily, calibration methods can unexpectedly degrade performance of \ct~(on \textsc{Msra}) due to inadequate hyperparameter search. Further, the comparison of Head and Tail scores hints that \pt~improves long-tailed performance by keeping a better trade-off between the head and tail, where \pt~achieves much better results on the tail. Yet, \pt~could slightly give a negative impact to Head scores.

Overall speaking, we can put that our hypothesis is valid, indicating a positive effect of \pt.

\subsection{Weight Norm Visualization}

\begin{figure}
    \centering
    \includegraphics[width=0.37\textwidth]{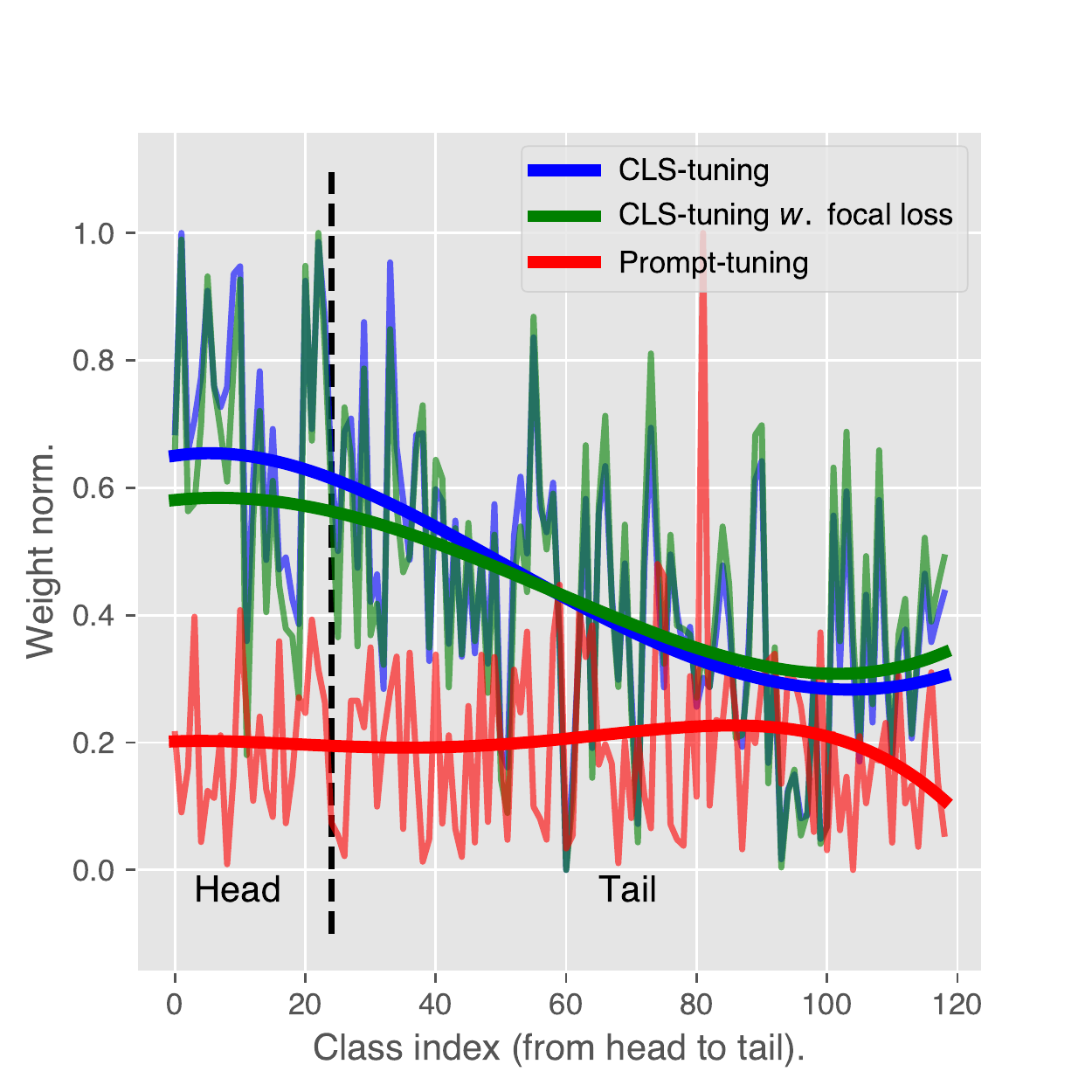}
    \caption{Weight norm visualization. Classification weight norms of the model trained on \textsc{Iflytek}.}
    \label{fig2}
\end{figure}

For sake of a deeper investigation of the hypothesis, we visualize the weight norms customized to different classes. The weights are essentially derived from the predictor of the classifier $\mathcal{C}$. The motivation behind the visualization is rooted on the widely accepted decoupling property~\citep{Kang20}, which claims that the learning of the backbone and classifier is in fact decoupled from each other. In other words, the long-tailed class distribution affects the classifier a lot, but might have little impact on the backbone. To this end, it is largely acknowledged in previous studies~\citep{Lin17,Kang20} that a good long-tailed learner should make weight norms roughly in the same scale, while a bad one sees a weight norm decay from the head to tail. The visualization is shown in Figure~\ref{fig2}.

From the plot, we discover that \ct~makes PLMs bad long-tailed learners, as it possesses large weight norms for the head while small ones for the tail. Applied to \ct, the focal loss slightly flattens the weight norm slope. As expected, \pt~makes PLMs own a way more flat distribution. This may be the reason why focal loss does not bring improvement over \pt. While one may wonder whether weight norm regularization could boost long-tailed performance, $\eta$-norm is actually a calibration method to adjust weight norms so that they can be in similar scales, showcasing minor improvement of weight norm regularization.

The contrast indicates that \pt~makes PLMs good long-tailed learners. 

\section{Analyses}

\label{sec4}

Although we have verified that our hypothesis is valid, we are more of thirst to see why \pt~can be so promising and provide further intuitions. To this demand, we consider three research questions here and carry out in-depth analyses to answer them. We also hope the analyses can shed light on the design of \pt~itself in related areas. 
\begin{itemize}[topsep=1pt,parsep=1pt]
    \item \textbf{RQ1}: Does the shared embedding contribute to \pt?
    \item \textbf{RQ2}: Does the input structure (i.e., MLM input) contribute to \pt?
    \item \textbf{RQ3}: Does the classifier structure and parameterization (e.g., layer normalization used in MLM head) contribute to \pt?
\end{itemize}

\begin{table*}[t]
    \centering
    \caption{Analysis results. \textsc{Avg} denotes average results over all datasets. The best \textsc{Avg} scores are boldfaced. The content after \ct~indicates the activation that is being used, where T is Tanh and R is ReLU. LN stands for layer normalization. pt. is short for pretrained and ed. is short for embedding decoupling. The variances are attached as subscripts.}
    \resizebox{1.0\textwidth}{!}{
    \begin{tabular}{lcccccccccccc}
    \toprule
    Dataset & \multicolumn{2}{c}{\textsc{Cmid}} & \multicolumn{2}{c}{\textsc{Iflytek}} & \multicolumn{2}{c}{\textsc{Ctc}} & \multicolumn{2}{c}{\textsc{Msra}} &  \multicolumn{2}{c}{\textsc{R52}} & \multicolumn{2}{c}{\textsc{Avg}} \\
    \midrule
    Metric & Acc & F1 & Acc & F1 & Acc & F1 & Acc & F1 & Acc & F1 & Acc & F1 \\
    \midrule
    \ct~$\circ$ T & 51.1\textsubscript{0.4} & 37.3\textsubscript{2.3} & 58.7\textsubscript{0.4} & 33.7\textsubscript{1.6} & 84.6\textsubscript{0.3} & 77.2\textsubscript{2.9} & 99.0\textsubscript{0.1} & 97.5\textsubscript{1.0} & 95.3\textsubscript{0.2} & 67.3\textsubscript{1.3} & 77.7 & 62.6 \\
    \quad w/ $\eta$-norm & 51.1\textsubscript{0.5} &	37.4\textsubscript{2.0} & 59.1\textsubscript{0.3} & 35.7\textsubscript{1.6} & 84.7\textsubscript{0.2} & 77.3\textsubscript{3.1} & 99.0\textsubscript{0.1} & 97.4\textsubscript{0.9} & 95.4\textsubscript{0.3} & 68.9\textsubscript{1.9} & 77.9 & 63.3 \\
    \quad w/ focal loss & 51.0\textsubscript{0.7} & 42.1\textsubscript{1.3} & 58.8\textsubscript{0.3} & 36.0\textsubscript{1.6} & 84.3\textsubscript{0.4} & 78.5\textsubscript{2.4} & 99.0\textsubscript{0.1} & 96.8\textsubscript{1.2} & 95.7\textsubscript{0.2} & 72.8\textsubscript{2.3} & 77.8 & 65.2 \\
    \midrule
    \ct~$\circ$ R & 50.9\textsubscript{0.4} & 34.5\textsubscript{1.4} & 58.7\textsubscript{0.3} & 33.3\textsubscript{1.1} & 84.4\textsubscript{0.4} & 77.1\textsubscript{1.0} & 99.0\textsubscript{0.1} & 97.7\textsubscript{0.5} & 94.2\textsubscript{0.4} & 56.2\textsubscript{2.5} & 77.4 & 59.8 \\
    \quad w/ $\eta$-norm & 50.9\textsubscript{0.5} & 34.8\textsubscript{1.8} & 58.4\textsubscript{0.3} & 33.3\textsubscript{1.0} & 84.6\textsubscript{0.4} & 78.0\textsubscript{1.5} & 99.1\textsubscript{0.0} & 97.8\textsubscript{0.5} & 94.3\textsubscript{0.3} & 56.3\textsubscript{1.9} & 77.5 & 60.0 \\
    \quad w/ focal loss & 51.0\textsubscript{0.5} & 40.1\textsubscript{1.5} & 58.8\textsubscript{0.4} & 34.6\textsubscript{0.1} & 84.6\textsubscript{0.3} & 76.9\textsubscript{0.6} & 99.0\textsubscript{0.1} & 97.0\textsubscript{1.5} & 95.1\textsubscript{0.3} & 66.0\textsubscript{2.7} & 77.7 & 62.9 \\
    \midrule
    \ct~$\circ$ R & \\
    \quad w/ prompt & 49.7\textsubscript{0.5} & 33.1\textsubscript{0.4} & 58.4\textsubscript{0.3} & 32.8\textsubscript{1.0} & 84.6\textsubscript{0.1} & 77.2\textsubscript{3.0} & 99.0\textsubscript{0.1} & 96.9\textsubscript{0.3} & 94.1\textsubscript{0.3} & 54.5\textsubscript{3.1} & 77.2 & 58.9 \\
    \quad w/ LN & 51.3\textsubscript{0.6} & 42.0\textsubscript{1.4} & 59.7\textsubscript{0.6} & 39.1\textsubscript{0.8} & 84.6\textsubscript{0.5} & 79.4\textsubscript{2.2} & 99.1\textsubscript{0.1} & 97.1\textsubscript{0.8} & 96.1\textsubscript{0.2} & 77.7\textsubscript{3.5} & \textbf{78.2} & 67.1 \\
    \quad w/ pt. LN & 50.8\textsubscript{0.6} & 42.5\textsubscript{1.2} & 59.4\textsubscript{0.4} & 41.4\textsubscript{0.9} & 84.4\textsubscript{0.5} & 79.7\textsubscript{1.5} & 99.1\textsubscript{0.1} & 97.7\textsubscript{0.5} & 96.2\textsubscript{0.2} & 82.0\textsubscript{1.8} & 78.0 & 68.7 \\
    \midrule
    \pt & 49.3\textsubscript{0.7} & 43.4\textsubscript{0.7} & 61.2\textsubscript{0.6} & 44.4\textsubscript{1.0} & 84.2\textsubscript{0.1} & 80.9\textsubscript{0.1} & 99.1\textsubscript{0.0} & 97.8\textsubscript{0.3} & 95.7\textsubscript{0.1} & 85.3\textsubscript{0.6} & 77.9 & \textbf{70.4} \\
    \quad w/ ed. & 49.4\textsubscript{0.7} & 43.6\textsubscript{0.7} & 61.0\textsubscript{0.7} & 44.4\textsubscript{1.0} & 84.2\textsubscript{0.4} & 80.5\textsubscript{0.9} & 99.0\textsubscript{0.2} & 96.9\textsubscript{1.4} & 95.7\textsubscript{0.2} & 84.9\textsubscript{1.0} & 77.9 & 70.1 \\
    \midrule
    Metric & Head & Tail & Head & Tail & Head & Tail & Head & Tail & Head & Tail & Head & Tail \\
    \midrule
    \ct~$\circ$ T & 50.3\textsubscript{1.0} & 34.1\textsubscript{3.0} & 61.8\textsubscript{0.6} & 27.4\textsubscript{1.9} & 87.7\textsubscript{0.2} & 74.1\textsubscript{3.7} & 99.2\textsubscript{0.1} & 97.4\textsubscript{1.1} & 99.0\textsubscript{0.1} & 66.6\textsubscript{1.3} & 79.6 & 59.9 \\
    \quad w/ $\eta$-norm & 50.3\textsubscript{0.9} & 34.3\textsubscript{2.7} & 62.1\textsubscript{0.4} & 29.7\textsubscript{2.0} & 87.8\textsubscript{0.2} & 74.3\textsubscript{4.0} & 99.2\textsubscript{0.1} & 97.3\textsubscript{1.0} & 99.0\textsubscript{0.1} & 68.3\textsubscript{1.9} & 79.7 & 60.8 \\
    \quad w/ focal loss & 49.8\textsubscript{0.8} & 40.2\textsubscript{1.5} & 62.0\textsubscript{0.4} & 30.2\textsubscript{1.9} & 87.5\textsubscript{0.3} & 75.9\textsubscript{3.1} & 99.3\textsubscript{0.0} & 96.7\textsubscript{1.3} & 99.0\textsubscript{0.0} & 72.3\textsubscript{2.4} & 79.5 & 63.1 \\
    \midrule
    \ct~$\circ$ R & 50.7\textsubscript{0.4} & 30.6\textsubscript{1.7} & 61.7\textsubscript{0.6} & 26.8\textsubscript{1.3} & 87.5\textsubscript{0.2} & 74.1\textsubscript{1.3} & 99.2\textsubscript{0.1} & 97.6\textsubscript{0.5} & 99.0\textsubscript{0.1} & 55.4\textsubscript{2.6} & 79.6 & 56.9 \\
    \quad w/ $\eta$-norm & 50.4\textsubscript{0.7} & 31.0\textsubscript{2.2} & 61.7\textsubscript{0.7} & 26.8\textsubscript{1.4} & 87.7\textsubscript{0.3} & 75.1\textsubscript{1.9} & 99.2\textsubscript{0.1} & 97.7\textsubscript{0.5} & 99.0\textsubscript{0.2} & 55.5\textsubscript{2.0} & 79.6 & 57.2 \\
    \quad w/ focal loss & 50.3\textsubscript{0.5} & 37.6\textsubscript{1.7} & 61.9\textsubscript{0.3} & 28.4\textsubscript{1.6} & 87.6\textsubscript{0.3} & 73.7\textsubscript{0.8} & 99.2\textsubscript{0.1} & 96.9\textsubscript{1.6} & 99.0\textsubscript{0.2} & 65.3\textsubscript{2.8} & 79.6 & 60.4 \\
    \midrule
    \ct~$\circ$ R & \\
    \quad w/ prompt & 49.7\textsubscript{0.9} & 29.1\textsubscript{0.5} & 61.5\textsubscript{0.9} & 26.3\textsubscript{1.3} & 87.5\textsubscript{0.1} & 74.2\textsubscript{3.9} & 99.2\textsubscript{0.1} & 96.8\textsubscript{0.3} & 99.0\textsubscript{0.1} & 53.6\textsubscript{3.2} & 79.4 & 56.0 \\
    \quad w/ LN & 50.0\textsubscript{0.7} & 40.0\textsubscript{1.6} & 63.1\textsubscript{0.6} & 33.7\textsubscript{0.9} & 87.6\textsubscript{0.4} & 77.0\textsubscript{2.7} & 99.3\textsubscript{0.2} & 97.0\textsubscript{0.8} & 99.0\textsubscript{0.1} & 77.3\textsubscript{3.6} & \textbf{79.8} & 65.0 \\
    \quad w/ pt. LN & 49.5\textsubscript{0.5} & 40.8\textsubscript{1.4} & 62.7\textsubscript{0.3} & 36.6\textsubscript{1.1} & 87.5\textsubscript{0.4} & 77.4\textsubscript{1.9} & 99.2\textsubscript{0.1} & 97.6\textsubscript{0.6} & 99.0\textsubscript{0.1} & 81.6\textsubscript{1.9} & 79.6 & 66.8 \\
    \midrule
    \pt & 48.4\textsubscript{1.0} & 42.2\textsubscript{0.7} & 63.6\textsubscript{0.4} & 40.1\textsubscript{1.2} & 	87.4\textsubscript{0.2} & 79.0\textsubscript{0.2} & 99.2\textsubscript{0.1} & 97.7\textsubscript{0.3} & 98.6\textsubscript{0.1} & 85.0\textsubscript{0.6} & 79.4 & \textbf{68.8} \\
    \quad w/ ed. & 48.2\textsubscript{1.0} & 42.5\textsubscript{0.7} & 63.6\textsubscript{0.4} & 40.1\textsubscript{1.2} & 87.4\textsubscript{0.3} & 78.5\textsubscript{1.1} & 99.3\textsubscript{0.1} & 96.8\textsubscript{1.5} & 98.7\textsubscript{0.2} & 84.6\textsubscript{1.0} & 79.4 & 68.5 \\
    \bottomrule
    \end{tabular}
    }
    \label{tab3}
\end{table*}

\subsection{Impact of Shared Embedding}

The first question comes into our mind is that whether it is the parameter sharing between the classifier and backbone that helps \pt~survive from collapsing. Hence, we decouple the parameters shared by the classifier and backbone (i.e., without shared embedding during optimization) and compare the results in Table~\ref{tab3} before and after the parameter decoupling.

We observe that the decoupling somehow has little impact on the performance (\pt~v.s. \pt~w/ ed.). Reversely, ed. even degrades the performance of \pt. The phenomenon gives \textit{a possibly negative response to} \textbf{RQ1}. 

\subsection{Impact of Input Structure}

We explore whether the input structure is a significant factor regarding long-tailed performance. To this end, we arm \ct~with MLM input so that \ct~may share the input structure and representation for classification as \pt.

The results in Table~\ref{tab3} demonstrate that the MLM input somewhat affects the performance of \ct, in terms of both Acc and F1 scores (\ct~v.s. \ct~w/ prompt). We attribute the performance detriment to mismatch of the CLS head and MLM input. That is, PLMs are not pretrained in the way that CLS head should decode the MLM input. And the results naturally suggest \textit{a possibly negative response to} \textbf{RQ2}. 

\subsection{Impact of Classifier Structure and Parameterization}

We also investigate the impact of the classifier structure and parameterization, given structural and parameter differences between classifier heads used by \ct~and \pt. We aim to check whether it is the discrepancy between classifiers that biases the learning. To study the impact, we replace the Tanh in \ct~with ReLU. We use ReLU here as an alternative of GELU owing to the fact that ReLU is more prevalent in the finetuning stage. Then, by adding a layer normalization after the ReLU activation, we fill the structural gap between two classifiers. We also perform a natural follow-on action, re-using the statistics of the layer normalization from the MLM head to further enhance the classifier. The results are presented in Table~\ref{tab3}. 

It is revealed that the ReLU variant sometimes yields surprisingly deteriorated results when compared to \ct~with Tanh. However, when the ReLU variant is additionally armed with a succeeding layer normalization (\ct~w/ LN), it can surpass the original \ct~by certain margins. Notably, \ct~w/ LN has better Acc scores than \pt~does, potentially suggesting the balanced use of \ct~w/ LN in real-world applications. Besides, by re-using the MLM layer normalization (\ct~w/ pt. LN), \ct~approximates \pt~at once. The results imply \textit{an absolutely positive response to} \textbf{RQ3}. We conjecture the observation is underpinned by an information perspective towards regulated features.

For the ReLU variant, negative features will be zero out, leading to a cut-down of information. The information cut-down can be referred to as ``dying ReLU'' problem~\citep{Lu19,DyReLU17} when negative features take a large portion. The information loss may be problematic to the head, and even more unfriendly to the tail that is under-represented (i.e., with much fewer examples). As a consequence, under-represented classes can be represented with high bias (and potentially high variance). In contrast, Tanh manipulates features without any tailoring, but restricts values to a constant range (i.e., from -1 to 1). Despite the reduced learning burden, Tanh suffers from large saturation area~\citep{XuHL16}. Thereby, the information of some tail classes can be compressed with detriment. Existing literature~\citep{GirshickDDM14,HeZRS15,XuWCL15} calibrates the situation by relaxing negative features. On the other hand, the layer normalization can compensate the information loss caused by ReLU. With learned affinities (a.k.a., element-wise weights and biases), the layer normalization shall properly re-locate and re-scale the ReLU-activated features so that knowledge attained from the head can be transferred to the tail and the debuff of ReLU can be alleviated. 


\begin{figure}
    \centering
    \includegraphics[width=0.49\textwidth]{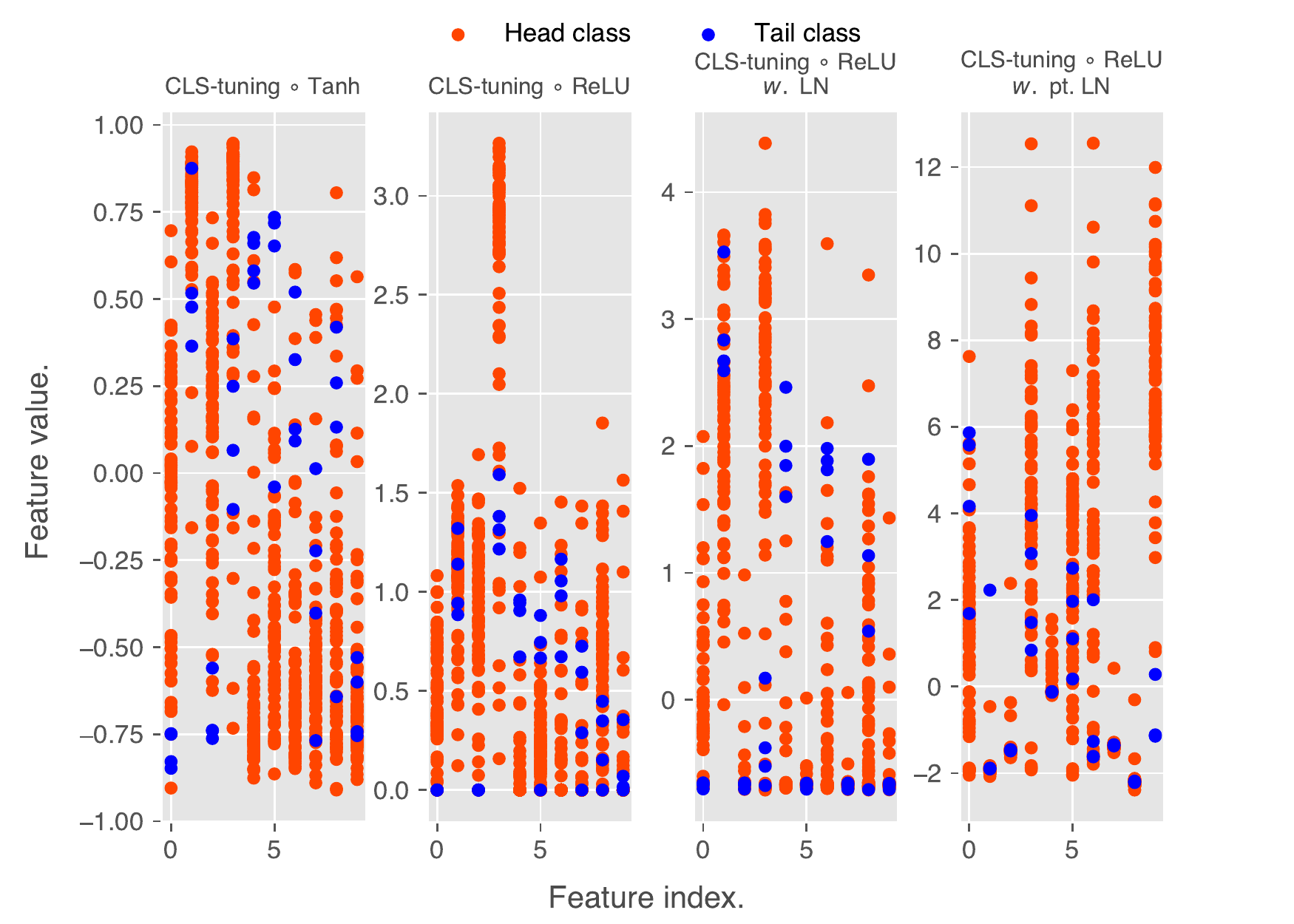}
    \caption{An illustration of sampled feature distributions, for both the head and tail.}
    \label{fig3}
\end{figure}

We add an illustration via Figure~\ref{fig3} for a more intuitive understanding of the above explanation. Taking examples of a sampled head class (\textit{commerce}) and a sampled tail class (\textit{delivery}) from \textsc{Iflytek}, we plot the distributions of the first 10 features in feature vectors (i.e., final hidden state vectors correspond to \texttt{[CLS]} tokens) derived from these examples. From the plot, we can see that ReLU certainly drops information of quite a few features of the tail (6 out of 10) and Tanh compulsively has features descend into the pre-defined range to involve themselves in optimization, both limiting the expressiveness of features. The randomly initialized layer normalization re-activates the dead features from the ReLU by re-locating them, but largely leaving them in fixed scales (6 out of 10). Furthermore, the pretrained layer normalization from the MLM head transfers knowledge from the head to tail and makes the features of the tail diversely distributed by re-scaling them.


    

\section{Applicability to Few-shot Classification}

From the analyses above, we come up with an alternative to \ct~that owns a comparable effectiveness with \pt~for long-tailed classification, by equipping \ct~with the structure and the layer normalization affinities of the pretrained MLM head. With this much simpler surrogate classifier, we prefer to retrospectively explore how the classifier would perform in the few-shot scenario, especially when compared with \pt. The intuition is that, while the layer normalization can not transfer knowledge from the head to tail any longer in few-shot classification, its innate friendliness to the tail can probably be applied to few-shot classes for generalization.

Therefore, we conduct experiments on three few-shot datasets, \textsc{Ecom} from FewCLUE~\citep{Xu21}, and \textsc{Rte}, \textsc{BoolQ} from SuperGLUE~\cite{Wang19b,SchickS21}. \textsc{Ecom} is a review sentiment classification dataset, \textsc{Rte}~\citep{Wang19a} is a two-way entailment dataset, and \textsc{BoolQ}~\citep{Clark19} is a yes-or-no question answering dataset.

\begin{table}
    \centering
    \caption{Statistics of few-shot datasets.}
    \resizebox{0.37\textwidth}{!}{
    \begin{tabular}{ccc}
    \toprule
    Dataset & \#Test exam. & \#Avg. tokens \\
    \midrule
    \textsc{Ecom} & 610 & 47.7 \\
    \textsc{Rte} & 277 & 52.3 \\
    \textsc{BoolQ} & 3,270 & 105.3 \\
    \bottomrule
    \end{tabular}
    }
    \label{tab4}
\end{table}

While \textsc{Ecom} has already been designed for few-shot learning, the others are not originally. For the latter two, we treat the original development set as the test set following~\citet{Gao21}, and randomly sample 32 examples uniformly from the original training set as the training set following~\citet{SchickS21}. To strictly following a \textit{true} few-shot setting~\citep{Perez21}, 32 examples that do not overlap with those in our training set are used to form the development set. The implementation is akin to the one we used for the long-tailed experiments. The batch size is reduced to 2 due to the small scale of training data. \texttt{bert-base-uncased} is leveraged as the backbone for English datasets. The maximum length is set to 64, 128, and 256 respectively for \textsc{Ecom}, \textsc{Rte}, and \textsc{BoolQ}. The statistics of these datasets are listed in Table~\ref{tab4}.

The example templates and verbalizers are listed in Appendix~\ref{app2}.

Moreover, since few-shot experiments are sensitive to the choice of hyperparameters, we again take average accuracy scores over 5 runs as the results, attached with variances.

\begin{table}
    \centering
    \caption{Applicability to few-shot classification results. \textsc{Avg} denotes average results over all datasets. The best \textsc{Avg} scores are boldfaced. The variances are attached as subscripts.}
    \resizebox{0.45\textwidth}{!}{
    \begin{tabular}{lcccc}
    \toprule
    Dataset & \textsc{Ecom} & \textsc{Rte} & \textsc{BoolQ} & \textsc{Avg} \\
    \midrule
    Metric & Acc & Acc & Acc & Acc \\
    \midrule
    \ct~$\circ$ T & 63.6\textsubscript{10.5} & 48.6\textsubscript{3.9} & 60.1\textsubscript{1.3} & 57.4 \\
    \midrule
    \ct~$\circ$ R & 64.7\textsubscript{2.7} & 48.7\textsubscript{3.6} & 55.2\textsubscript{7.5} & 56.2 \\
    \quad w/ prompt & 60.0\textsubscript{5.5} & 49.7\textsubscript{2.3} & 59.9\textsubscript{4.3} & 56.5 \\
    \quad w/ pt. LN & 68.7\textsubscript{5.0} & 55.0\textsubscript{1.8} & 59.1\textsubscript{1.7} & 60.9 \\
    \midrule
    \pt & 77.3\textsubscript{9.3} & 51.8\textsubscript{1.8} & 60.7\textsubscript{1.0} & \textbf{63.3} \\
    \bottomrule
    \end{tabular}
    }
    \label{tab5}
\end{table}

We can observe from Table~\ref{tab5} that \ct~with ReLU and pretrained layer normalization can perform better than other \ct~baselines, but may not necessarily hold across datasets. The filled gap between \ct~and \pt~shows the scalability of our finding to few-shot classification. However, we encourage future work to explore the regime for a more comprehensive understanding.

\section{Related Work}

PLMs have brought classification tasks to a brand-new stage where the solutions to these tasks are way much simpler than ever~\cite{Devlin19,Wang19a}. However, PLMs are still sub-optimal for some corner cases, such as few-shot classification~\citep{Zhang21} and long-tailed classification~\citep{Li20}.

For few-shot classification, \pt, which finetunes models in a language modeling fashion, is increasingly taking a central role in the mainstream methods~\citep{Liu21}. Since PLMs are mostly trained with language modeling objectives, \pt~becomes, as water is to fish, the key to unearthing the few-shot or even zero-shot learning capabilities of PLMs~\citep{Brown20,SchickS21,Scao21}. Instantly after the success, the engineering of prompts drives related research on prompt search/generation~\citep{Jiang20,Shin20,Gao21} and various downstream applications such as text generation~\citep{LiX20}, relation extraction~\citep{Chen20}, and entity typing~\citep{Ding2021}. Recently, \pt~also serves as an alternative way for parameter-efficient finetuning by only finetuning parameters of the inserted continuous prompts~\citep{Lester21}, in place of previously adopted adapter-tuning~\citep{Houlsby19}. The parameter efficiency brought by \pt~has blazed a trail for increasingly large language models.

In contrast, little work has been investigated to make PLMs good long-tailed learners. Intuitively, the tail classes are essentially few-shot ones. Thus, we speculate that \pt~is also a promising choice to make PLMs good long-tailed learners via transferring knowledge of head classes. As long-tailed classification is a long-standing problem in the general area of machine learning~\citep{Lin17,LiuZ19,Zhou20,Kang20,Tang20} and the long-tailed phenomenon also exists in the domain of natural language processing, we believe that a systematic exploration on whether and why \pt~ can make PLMs good long-tailed learners will facilitate further advances in the related areas.

\section{Conclusions}

Inspired by the success of \pt~in few-shot learning, we empirically examine whether \pt~can make PLMs good long-tailed learners in this work. The results validate the hypothesis. We also conduct in-depth analyses on why \pt~benefits PLMs for long-tailed classification, from the perspectives of coupling, classifier, and input respectively, to offer further intuitions. Based on the analyses, we summarize that the classifier structure and parameterization are crucial for enhancing long-tailed performance of PLMs, in contrast to other factors. Extended empirical evaluation results on few-shot classification show that our finding shed light on related work that seeks to boost \pt.

\section*{Limitations}

Since prompt-tuning is shown sensitive to small variations of templates, prompt-tuning should be performed with reasonable templates. However, we do not study the impact of different templates since our work is not concerned with finding a good template for long-tailed classification.


\section*{Acknowledgements}

We would like to thank Yujia Qin and Zhiyuan Liu from Tsinghua University for their helpful suggestions. 

\bibliography{anthology,custom}
\bibliographystyle{acl_natbib}

\clearpage
\appendix

\section{Templates and Verbalizers for Long-tailed Datasets}
\label{app1}

The templates and verbalizers for three long-tailed datasets are separately listed in Table~\ref{tab6}, Table~\ref{tab7}, Table~\ref{tab8}, Table~\ref{tab9}, and Table~\ref{tab10}.

\begin{table*}[t]
    \centering
    \resizebox{0.83\textwidth}{!}{
    \begin{tabular}{cc}
    \toprule
    Dataset & \textsc{Cmid} \\
    \midrule
    Template & \zh{$x$?这个问题的意图是\texttt{[MASK]}.} \\
    \midrule
    Verbalizer & \zh{\makecell[c]{
    label:病症治疗方法$\to$病症治疗方法(disease treatment),\\
    label:病症定义$\to$病症定义(disease definition),\\
    label:病症临床表现(病症表现)$\to$病症临床表现病症表现(disease symptom),\\
    label:药物适用症$\to$药物适用症(medicine applicability),\\
    label:其他无法确定$\to$其他无法确定(others undefined),\\
    label:病症禁忌$\to$病症禁忌(disease contradiction),\\
    label:病症相关病症$\to$病症相关病症(disease related diseases),\\
    label:其他对比$\to$其他对比(others contrast),\\
    label:药物副作用$\to$药物副作用(medicine side-effect),\\
    label:药物禁忌$\to$药物禁忌(medicine contradiction),\\
    label:其他多问$\to$其他多问(others multiple questions),\\
    label:病症病因$\to$病症病因(disease cause),\\
    label:治疗方案化验/体检方案$\to$治疗方案化验体检方案(treatment exmination),\\
    label:治疗方案恢复$\to$治疗方案恢复(treatment recovery),\\
    label:病症严重性$\to$病症严重性(disease severity),\\
    label:病症治愈率$\to$病症治愈率(disease cure rate),\\
    label:药物用法$\to$药物用法(medicine usage),\\
    label:药物作用$\to$药物作用(medicine effect),\\
    label:其他两性$\to$其他两性(others sex),\\
    label:治疗方案正常指标$\to$治疗方案正常指标(treatment normal indication),\\
    label:其他养生$\to$其他养生(others health),\\
    label:治疗方案方法$\to$治疗方案方法(treatment method),\\
    label:病症传染性$\to$病症传染性(disease infectivity),\\
    label:药物成分$\to$药物成分(medicine component),\\
    label:病症预防$\to$病症预防(disease prevention),\\
    label:治疗方案恢复时间$\to$治疗方案恢复时间(treatment recovery duration),\\
    label:病症推荐医院$\to$病症推荐医院(disease recommende hospital),\\
    label:治疗方案费用$\to$治疗方案费用(treatment cost),\\
    label:治疗方案临床意义$\to$治疗方案临床意义(treatment significance),\\
    label:其他设备用法$\to$其他设备用法(others device usage),\\
    label:治疗方案疗效$\to$治疗方案疗效(treatment efficacy),\\
    label:药物价钱$\to$药物价钱(medicine price),\\
    label:治疗方案有效时间$\to$治疗方案有效时间(treatment effective duration),\\
    label:其他整容$\to$其他整容(others cosmetic),\\
    label:病症所属科室$\to$病症所属科室(disease department),\\
    label:治疗方案治疗时间$\to$治疗方案治疗时间(treatment duration)
    }} \\
    \bottomrule
    \end{tabular}
    }
    \caption{Template and verbalizer for \textsc{Cmid} (with necessary translations).}
    \label{tab6}
\end{table*}

\begin{table*}[t]
    \centering
    \resizebox{0.97\textwidth}{!}{
    \begin{tabular}{cc}
    \toprule
    Dataset & \textsc{Iflytek} \\
    \midrule
    Template & \zh{$x$.这句话描述的物品属于\texttt{[MASK]}.} \\
    \midrule
    Verbalizer & \zh{\makecell[c]{
    label:打车$\to$打车(taxi),label:导航$\to$导航(navigation),label:WIFI$\to$wifi,\\
    label:租车$\to$租车(car rent),label:同城$\to$同城(urban service),label:快递$\to$快递(express),\\
    label:婚庆$\to$婚庆(wedding),label:家政$\to$家政(house service),\\
    label:公共交通$\to$公共交通(public transport),label:政务$\to$政务(government affair),\\
    label:社区服务$\to$社区服务(community service),label:薅羊毛$\to$薅羊毛(deal hunter),\\
    label:魔幻$\to$魔幻(magic game),label:仙侠$\to$仙侠(warrior game),\\
    label:卡牌$\to$卡牌(card game),label:空战$\to$空战(flight game),\\
    label:射击$\to$射击(shooting game),label:休闲$\to$休闲(leisure game),\\
    label:动作$\to$动作(action game),label:体育$\to$体育(sports game),\\
    label:棋牌$\to$棋牌(board game),label:养成$\to$养成(simulation game),\\
    label:策略$\to$策略(strategy game),label:MOBA$\to$MOBA,label:辅助工具$\to$辅助工具(aid tool),\\
    label:约会$\to$约会(date),label:通讯$\to$通讯(communication),label:工作$\to$工作(work),\\
    label:论坛$\to$论坛(BBS),label:婚恋$\to$婚恋(marriage),label:情侣$\to$情侣(lover),\\
    label:社交$\to$社交(social),label:生活$\to$生活(life),label:博客$\to$博客(blog),\\
    label:新闻$\to$新闻(news),label:漫画$\to$漫画(cartoon),
    label:小说$\to$小说(novel),\\
    label:技术$\to$技术(technology),label:教辅$\to$教辅(teaching),label:问答$\to$问答(QA),\\
    label:搞笑$\to$搞笑(fun),label:杂志$\to$杂志(magazine),label:百科$\to$百科(wikipedia),\\
    label:影视$\to$影视(TV),label:求职$\to$求职(job),label:兼职$\to$兼职(part-time),\\
    label:视频$\to$视频(video),label:短视频$\to$短视频(clips),label:音乐$\to$音乐(music),\\
    label:直播$\to$直播(live),label:电台$\to$电台(radio),label:K歌$\to$k歌(KTV),\\
    label:成人$\to$成人(adult),label:中小学$\to$中小学(school),label:职考$\to$职考(exam),\\
    label:公务员$\to$公务员(civil servant),label:视频教育$\to$视频教育(video edu.),\\
    label:高等教育$\to$高等教育(advanced edu.),label:成人教育$\to$成人教育(adult edu.),\\
    label:艺术$\to$艺术(art),label:语言(非英语)$\to$语言非英语(non-english),\\
    label:英语$\to$英语(english),label:旅游$\to$旅游(travel),label:预定$\to$预定(preservation),\\
    label:民航$\to$民航(flight),label:铁路$\to$铁路(railway),label:酒店$\to$酒店(hotel),\\
    label:行程$\to$行程(route),label:民宿$\to$民宿(BnB),label:出国$\to$出国(abroad),\\
    label:工具$\to$工具(general tool),label:亲子$\to$亲子(kid),label:母婴$\to$母婴(infant),\\
    label:驾校$\to$驾校(driver),label:违章$\to$违章(violation),label:汽车$\to$汽车(vehicle),\\
    label:汽车交易$\to$汽车交易(vehicle trade),label:养车$\to$养车(vehicle maintainance),\\
    label:行车$\to$行车(driving aid),label:租房$\to$租房(house rent),\\
    label:买房$\to$买房(house purchase),label:装修$\to$装修(house decoration),\\
    label:问诊$\to$问诊(inquiry),label:养生$\to$养生(health preservation),\\
    label:医疗$\to$医疗(medical),label:减肥$\to$减肥(slim),label:美妆$\to$美妆(makeup),\\
    label:菜谱$\to$菜谱(menu),label:餐饮$\to$餐饮(restaurant),\\
    label:健身$\to$健身(fit),label:支付$\to$支付(payment),label:保险$\to$保险(insurance),\\
    label:股票$\to$股票(stock),label:借贷$\to$借贷(loan),label:理财$\to$理财(finance management),\\
    label:彩票$\to$彩票(lottery),label:记账$\to$记账(acount keeping),label:银行$\to$银行(bank),\\
    label:剪辑$\to$剪辑(film editing),label:修图$\to$修图(photo editing),\\
    label:相机$\to$相机(camera),label:绘画$\to$绘画(painting),label:二手$\to$二手(second-hand),\\
    label:电商$\to$电商(e-commerce),label:团购$\to$团购(group purchase),\\
    label:外卖$\to$外卖(take-out),label:美颜$\to$美颜(beauty),label:电子$\to$电子(electronics),\\
    label:电影$\to$电影(movies),label:超市$\to$超市(market),label:购物$\to$购物(shopping),\\
    label:笔记$\to$笔记(notes),label:办公$\to$办公(office),label:日程$\to$日程(schedule),\\
    label:女性$\to$女性(female),label:经营$\to$经营(management),label:收款$\to$收款(check),\\
    label:体育咨讯$\to$体育咨讯(sports info.),label:其他$\to$其他(others)
    }} \\
    \bottomrule
    \end{tabular}
    }
    \caption{Template and verbalizer for \textsc{Iflytek} (with necessary translations).}
    \label{tab7}
\end{table*}

\begin{table*}[t]
    \centering
    \resizebox{0.83\textwidth}{!}{
    \begin{tabular}{cc}
    \toprule
    Dataset & \textsc{Ctc} \\
    \midrule
    Template & \zh{$x$.这个标准的类别是\texttt{[MASK]}.} \\
    \midrule
    Verbalizer & \zh{\makecell[c]{
    label:治疗或手术$\to$治疗或手术(therapy or surgery),\\
    label:体征$\to$体征(body sign),label:成瘾行为$\to$成瘾行为(addictive behavior),\\
    label:年龄$\to$年龄(age),label:疾病$\to$疾病(disease),\\
    label:器官组织状态$\to$器官组织状态(organ or tissue status),\\
    label:过敏耐受$\to$过敏耐受(allergy tolerance),\\
    label:依存性$\to$依存性(compliance with protocol),\\
    label:风险评估$\to$风险评估(risk assessment),\\
    label:怀孕相关$\to$怀孕相关(pregnancy-related activity),\\
    label:诊断$\to$诊断(diagnostic),label:复合$\to$复合(multiple criteria),\\
    label:实验室检查$\to$实验室检查(laboratory examination),\\
    label:知情同意$\to$知情同意(consent),label:献血$\to$献血(blood donation),\\
    label:参与其它试验$\to$参与其它试验(enrollment in other studies),\\
    label:药物$\to$药物(pharmaceutical substance or drug),\\
    label:能力$\to$能力(capacity),label:饮食$\to$饮食(diet),\\
    label:特殊病人体征$\to$特殊病人体征(special patient characteristic),\\
    label:疾病分期$\to$疾病分期(non-Neoplasm disease stage),\\
    label:研究者决定$\to$研究者决定(researcher decision),\\
    label:数据可及性$\to$数据可及性(data accessiblility),\\
    label:预期寿命$\to$预期寿命(life expectancy),\\
    label:肿瘤进展$\to$肿瘤进展(neoplasm status),\\
    label:读写能力$\to$读写能力(literacy),\\
    label:病例来源$\to$病例来源(patience source),\\
    label:锻炼$\to$锻炼(exercise),label:症状$\to$症状(symptom),\\
    label:受体状态$\to$受体状态(receptor status),\\
    label:口腔相关$\to$口腔相关(oral related),\\
    label:种族$\to$种族(ethnicity),\\
    label:健康群体$\to$健康群体(the healthy),\\
    label:残疾群体$\to$残疾群体(disabilities),\\
    label:设备$\to$设备(device),label:性别$\to$性别(gender),\\
    label:吸烟状况$\to$吸烟状况(smoking status),\\
    label:性取向$\to$性取向(sex related),\\
    label:护理$\to$护理(nursing),label:睡眠$\to$睡眠(bedtime),\\
    label:酒精使用$\to$酒精使用(alcohol consumption),\\
    label:居住情况$\to$居住情况(living condition),\\
    label:教育状况$\to$教育状况(education),\\
    label:伦理审查$\to$伦理审查(ethical audit)
    }} \\
    \bottomrule
    \end{tabular}
    }
    \caption{Template and verbalizer for \textsc{Ctc} (with necessary translations).}
    \label{tab8}
\end{table*}

\begin{table*}[t]
    \centering
    \resizebox{0.87\textwidth}{!}{
    \begin{tabular}{cc}
    \toprule
    Dataset & \textsc{Msra} \\
    \midrule
    Template & \zh{$x$.这句话中的$e$是\texttt{[MASK]}.} \\
    \midrule
    Verbalizer & \zh{\makecell[c]{
    label:integer$\to$整数(integer),label:ordinal$\to$序数(ordinal),\\
    label:location$\to$地点(location),label:date$\to$日期(date),\\
    label:organization$\to$机构(organization),label:person$\to$人物(person),\\
    label:money$\to$钱款(money),label:duration$\to$时段(duration),\\
    label:time$\to$时间(time),label:length$\to$长度(length),\\
    label:age$\to$年龄(age),label:frequency$\to$频率(frequency),\\
    label:angle$\to$角度(angle),label:phone$\to$电话(phone),\\
    label:percent$\to$百分数(percent),label:fraction$\to$分数(fraction),\\
    label:weight$\to$重量(weight),label:area$\to$面积(area),\\
    label:capacity$\to$容积(capacity),label:decimal$\to$小数(decimal),\\
    label:measure$\to$其他度量(other measure),label:speed$\to$速度(speed),\\
    label:temperature$\to$温度(temperature),label:postal code$\to$邮政编码(postal code),\\
    label:rate$\to$比率(rate),label:www$\to$网址(website)                 
    }} \\
    \bottomrule
    \end{tabular}
    }
    \caption{Template and verbalizer for \textsc{Msra}.}
    \label{tab9}
\end{table*}

\begin{table*}[t]
    \centering
    \resizebox{0.75\textwidth}{!}{
    \begin{tabular}{cc}
    \toprule
    Dataset & \textsc{R52} \\
    \midrule
    Template & $x$. This is \texttt{[MASK]}. \\
    \midrule
    Verbalizer & \zh{\makecell[c]{
    label:copper$\to$copper,label:livestock$\to$livestock,\\
    label:gold$\to$gold,label:money-fx$\to$money fx,\\
    label:tea$\to$tea,label:ipi$\to$ipi,\\
    label:trade$\to$trade,label:cocoa$\to$cocoa,\\
    label:iron-steel$\to$iron steel,label:reserves$\to$reserves,\\
    label:zinc$\to$zinc,label:nickel$\to$nickel,\\
    label:ship$\to$ship,label:cotton$\to$cotton,\\
    label:platinum$\to$platinum,label:alum$\to$alum,\\
    label:strategic-metal$\to$strategic metal,label:instal-debt$\to$instal debt,\\
    label:lead$\to$lead,label:housing$\to$housing,\\
    label:gnp$\to$gnp,label:sugar$\to$sugar,\\
    label:rubber$\to$rubber,label:dlr$\to$dlr,\\
    label:tin$\to$tin,label:interest$\to$interest,\\
    label:income$\to$income,label:crude$\to$crude,\\   
    label:coffee$\to$coffee,label:jobs$\to$jobs,\\   
    label:meal-feed$\to$meal feed,label:lei$\to$lei,\\   
    label:lumber$\to$lumber,label:gas$\to$gas,\\   
    label:nat-gas$\to$nat gas,label:veg-oil$\to$veg oil,\\   
    label:orange$\to$orange,label:heat$\to$heat,\\   
    label:wpi$\to$wpi,label:cpi$\to$cpi,\\   
    label:earn$\to$earn,label:jet$\to$jet,\\   
    label:potato$\to$potato,label:bop$\to$bop,\\  
    label:money-supply$\to$money supply,label:carcass$\to$carcass,\\  
    label:acq$\to$acq,label:pet-chem$\to$pet chem,\\  
    label:grain$\to$grain,label:fuel$\to$fuel,\\  
    label:retail$\to$retail,label:cpu$\to$cpu,\\  
    }} \\
    \bottomrule
    \end{tabular}
    }
    \caption{Template and verbalizer for \textsc{R52}.}
    \label{tab10}
\end{table*}

\section{Templates and Verbalizers for Few-shot Datasets}
\label{app2}

The templates are:
\begin{itemize}[topsep=1pt,parsep=1pt]
    \item \textsc{Ecom}: \zh{$x$. 这个评论中的物品\texttt{[MASK]}好.} ($x$. The product mentioned in the review is \texttt{[MASK]} good.)
    \item \textsc{Rte} \& \textsc{BoolQ}: $p$. Question: $q$? The answer: \texttt{[MASK]}.
\end{itemize}
where $p$ denotes the passage or premise, and $q$ denotes the question or hypothesis. Accordingly, the verbalizers are:
\begin{itemize}[topsep=1pt,parsep=1pt]
    \item \textsc{Ecom}: \zh{\{label:positive$\to$很(very),\\label:negative$\to$不(not)\}}
    \item \textsc{Rte}: \{label:entailment$\to$yes,\\label:not\_entailment$\to$no\}
    \item \textsc{BoolQ}: \{label:true$\to$yes,label:false$\to$no\}
\end{itemize}
Here, English translations are additionally used for \textsc{Ecom}.

\end{document}